\documentclass{article} 
\usepackage{iclr_r2fm,times}

\usepackage{amsmath,amsfonts,bm}

\def\eqref#1{equation~\ref{#1}}

\def\1{\bm{1}}

\DeclareMathAlphabet{\mathsfit}{\encodingdefault}{\sfdefault}{m}{sl}
\SetMathAlphabet{\mathsfit}{bold}{\encodingdefault}{\sfdefault}{bx}{n}

\usepackage{hyperref}
\usepackage{url}
\usepackage{booktabs}
\usepackage{array,multirow,graphicx}
\usepackage{natbib}
\usepackage{xspace}

\usepackage[capitalize,noabbrev]{cleveref}
\hypersetup{
colorlinks=true,linkcolor=magenta,citecolor=teal
}

\newcommand{\method}{\textsc{LoCo-LM}\xspace}
\newcommand{\methods}{\textsc{LoCo-LMs}\xspace}

\title{Towards Logically Consistent Language Models via Probabilistic Reasoning}

\iclrfinalcopy

\author{
  Diego Calanzone
  \raisebox{3pt}{\scalebox{.2}{\includegraphics[width=.1\textwidth]{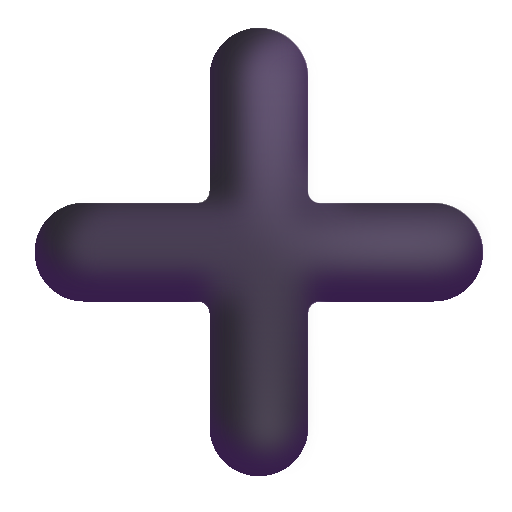}}}
  \thanks{Correspondence to \texttt{\{diego.calanzone\}@studenti.unitn.it}.  {{\scalebox{.2}{\includegraphics[width=.1\textwidth]{plus.png}}} $=$ University of Trento,
  {\scalebox{.2}{\includegraphics[width=.1\textwidth]{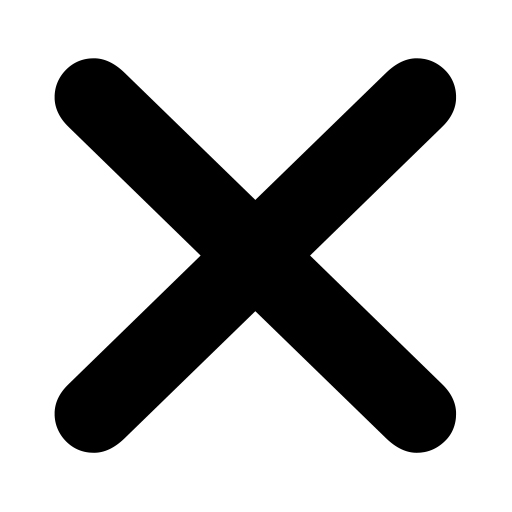}}} $=$ University of Edinburgh
  }, {\scalebox{.25}{\includegraphics[width=.1\textwidth]{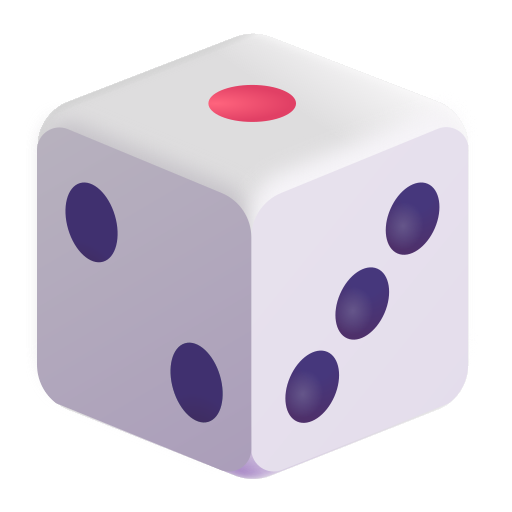}}} $=$ joint supervision.
  }
  \\
  \And
  Stefano Teso\raisebox{3pt}{\scalebox{.2}{\includegraphics[width=.1\textwidth]{plus.png}}}
  \raisebox{3pt}{\scalebox{.25}{\includegraphics[width=.1\textwidth]{dice.png}}}
  \\
  \And
  Antonio Vergari\raisebox{3pt}{\scalebox{.2}{\includegraphics[width=.1\textwidth]{prod.jpg}}}\raisebox{3pt}{\scalebox{.25}{\includegraphics[width=.1\textwidth]{dice.png}}}
  \\
}

\begin{document}

\maketitle

\begin{abstract}
Large language models (LLMs) are a promising venue for natural language understanding and generation tasks.
However, current LLMs are far from reliable: they are prone to generate non-factual information and,
more crucially, to contradict themselves when prompted to reason about beliefs of the world.
These problems are currently addressed with large scale fine-tuning or by delegating consistent reasoning to external tools.
In this work, we strive for a middle ground and introduce a training objective based on principled probabilistic reasoning that teaches a LLM to be consistent with external knowledge in the form of a set of facts and rules.
Fine-tuning with our loss on a limited set of facts enables our LLMs to be more logically consistent than previous baselines and allows them to extrapolate to unseen but semantically similar factual knowledge more systematically.
\end{abstract}

\section{Introduction}

Developing reliable large language models (LLMs) and safely deploying them is more and more crucial, particularly when they are used as external sources of knowledge \citep{petroni2019language}.
To do so, one would need LLMs to be \textit{factual} \citep{wadden2020fact}, i.e., generating content that satisfies some knowledge base (KB), and \textit{logically self-consistent} \citep{li2019logicdriven}, i.e., being able not to contradict themselves when prompted to perform complex reasoning.
Clearly, training on large datasets for question answering (QA) \citep{tafjord2021generalpurpose} alone cannot meet these desiderata \citep{evans2021truthful, lin2022truthfulqa, liu2023vera}.

Factuality and consistency are intimately related.
Enforcing factuality alone generally boils down to fine-tuning an LLM on a large KB of atomic facts \citep{kassner2021beliefbank}.
When predicting the truth values of these facts, a number of works try to enforce the simplest form of consistency: that the probability of a true fact shall be one minus the probability of its negation \citep{burns2022discovering}.
\citet{liu2023vera} use more sophisticated heuristics, fine-tuning on a large QA dataset with a combination of three objectives: binary cross entropy to classify true facts; multiclass cross entropy to choose the true statement among other options; a supervised contrastive loss to pull apart true and false facts.
All these approaches require large KBs and extrapolating to unseen facts remains an open challenge.

When it comes to self-consistency w.r.t. more complex reasoning scenarios, e.g., ensuring that LLMs can perform modus ponens without contradicting themselves \citep{tafjord2022entailer}, one line of research focuses on employing external reasoning tools such as MAX-SAT solvers \citep{Battiti2009} at inference time \citep{mitchell2022enhancing, jung2022maieutic, kassner2023language}.
However, these approaches depend on the constant availability of a reasoner (and sometimes also of a natural language inference model \citep{mitchell2022enhancing}) which can increase the cost of inference for every reasoning step.
At the same time, training the LLM to reason is not possible or hindered by the hardness of backpropagating through the solvers \citep{pogancicPMMR20}.

In this work, we show how to improve factuality and self-consistency of LLMs without external components by leveraging recent advancements in neuro-symbolic learning \citep{de2021statistical}.
This is done by turning complex reasoning tasks into logical constraints that can be incorporated in a semantic loss \citep{xu2018semantic}.
This in turn enforces the LLM to perform principled probabilistic reasoning at training time over the possible truth assignments.
Among previous works of research, \citet{zhang2023improved} similarly applied the semantic loss to instill integrity constraints in the embedding space of entities in encoder-only models.
It is worth noting that representations of beliefs can be highly contextual,
as shown by \citet{arakelyan2024semantic} with NLI models, suggesting additional forms of consistency to be included.
We empirically show how given incomplete factual knowledge, the LLM can learn truth beliefs for new facts while keeping logical consistency with prior knowledge.
In our experiments, with a single offline training session, LLMs trained with our objective outperform models relying on external solvers, and are more factual and logically consistent in low-data regimes when compared to standard supervised fine-tuning.

\begin{figure}[!t]
    \begin{center}
        \includegraphics[width=1\textwidth]{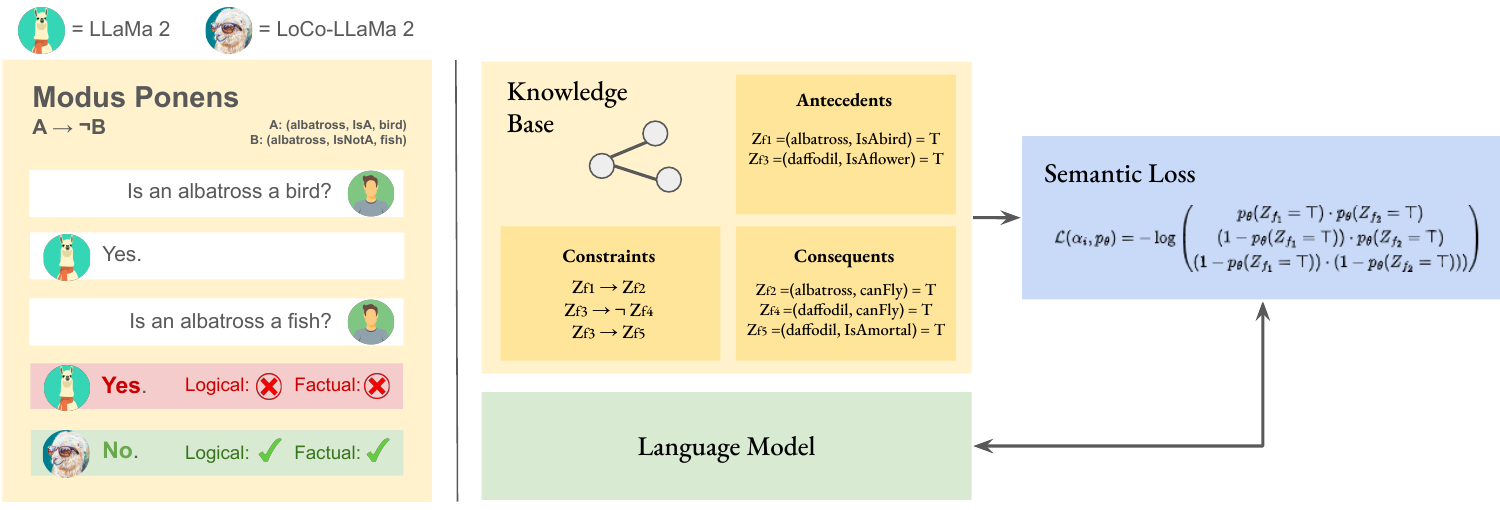}
    \end{center}
    \caption{Our \methods are trained by compiling implication constraints in a semantic loss and applying them to some atomic facts as antecedents. At test time, they can predict the missing consequents more consistently and factually than other baselines (\cref{table_A}).}
    \label{pipeline}
\end{figure}

\section{Tackling Factual \& Consistent Inference with \methods}
\label{sec:methodology}

\paragraph{Setting.}
We view a pre-trained LLM as a collection of truth beliefs (facts) over which it can \textit{factually reason}:  when prompted with question like ``Is a bison a mammal?", it can supply a binary prediction of the form ``Yes"/``No" or ``True"/``False".
When
given a limited set of textual statements $\mathcal{D}_f = \{f_1, \ldots, f_n\}$, such as ``an albatross is a bird'', and general logical implications (from here on, constraints) $\mathcal{D}_c = \{\alpha_1,\ldots,\alpha_m\}$, such as ``$\text{IsAbird} \rightarrow \text{canFly}$'',
we want an LLM queried about new beliefs $\{f_{n+1}, \ldots, f_k\}$ to output truth values that are \textit{consistent with the known facts} $\mathcal{D}_f$ and the prior KB $\mathcal{D}_c$.
We assume the facts are provided as \textit{(subject, property)} pairs, and that the logical implications specify how new properties (consequents) follow from the initial ones (antecedents).

\textbf{Logical constraints satisfaction}.
LLMs fit only on facts -- via, e.g., a cross-entropy loss -- struggle to infer consistent truth values for unseen consequents that satisfy the KB $\mathcal{D}_c$.
To design logically-consistent LLMs (\methods), we instead minimize a semantic loss (SL) \citep{xu2018semantic} that penalizes a model proportionally to the probability it allocates to truth values inconsistent with one or more such logical constraints.
To build intuition, consider two facts $f_1 = \text{(daffodil, IsAflower)}, f_2 = \text{(daffodil, IsMortal)}$ and a logical implication $\alpha_i$ stating that IsAflower entails IsMortal, or
\begin{equation}
    \label{example_constraint}
    \alpha_i := \; (z_{f_1} \rightarrow z_{f_2}) \iff \; (\lnot z_{f_1} \lor z_{f_2})
\end{equation}
where $z_f$ is the truth value assigned to a fact $f$.
The SL in \methods will penalize the LLM distribution $p_\theta$  if it allocates truth values inconsistent with $\alpha_i$, specifically $(z_{f_1} = \top, z_{f_2} = \bot)$ which encodes the false belief that daffodils are flowers ($\top =$ true) but not mortal ($\bot =$ false).
To this end, it relies on probabilities for these truth values extracted from the model.
We obtain these probabilities directly by reading off the likelihood of utterances produced by the LLM, that is:
\begin{align}
    p_\theta(z_{f_1} = \top )
        & = p_\theta(x_t = l \mid x_1,\ldots,x_{t-1} = \text{``Is a daffodil a flower?"})
    \\
    p_\theta(z_{f_2} = \top )
        & = p_\theta(x_t = l \mid x_1,\ldots,x_{t-1} = \text{``Is a daffodil a mortal?"})
\end{align}
Here, $l$ is a textual truth value, e.g., ``Yes" for $\top$ or ``No'' for $\bot$.
More generally, the SL encourages consistency by maximizing the likelihood of assignments satisfying $\alpha_i\in\mathcal{D}_c$, and can be applied to more complex reasoning scenarios that involve not only implications but any logical constraint.
From a constraint $\alpha_i \in \mathcal{D}_c$,  a \method is fine-tuned by minimizing a loss that is defined as:
\begin{equation}
    \label{semantic_loss}
    \mathcal{L}(\alpha_i, p_{\theta}) =
    - \log \sum\nolimits_{{\bf z} \models \alpha_i}\;\;\prod\nolimits_{j:{\bf z} \models z_{f_j}} p_\theta(z_{f_j})\;\;\prod\nolimits_{j:{\bf z} \models \lnot z_{f_j}} 1-p_\theta(z_{f_j})
\end{equation}
where $\mathrm{z} \models \alpha_i$ means that a set of truth assignment to facts satisfies $\alpha_i$ and the argument of the logarithm computes the probability of $\alpha_i$ being true.
Taking for instance the constraint in \cref{example_constraint}, the assignments from the truth table satisfying the formula are $\{(\top, \top), (\bot, \top), (\bot, \bot)\}$, the semantic loss in a \method would thus minimize the negative log likelihood of these assignments:
\begin{equation}
    \label{applied_semantic_loss_1}
    \begin{split}
        \mathcal{L}(\alpha_i,p_\theta) = - \log \begin{pmatrix}
            p_\theta(z_{f_1} = \top) \cdot p_\theta(z_{f_2} = \top) \;\; + \\
                (1 - p_\theta(z_{f_1} = \top)) \cdot p_\theta(z_{f_2} = \top) \;\; + \\
                (1 - p_\theta(z_{f_1} = \top)) \cdot (1 - p_\theta(z_{f_2} = \top)) \end{pmatrix}
    \end{split}
\end{equation}
\cref{pipeline} illustrates the pipeline of \methods.

\textbf{Factuality}.
As described so far, the SL in \methods fosters self-consistency, but can hinder factuality as it considers as equally valid assignments both the cases where antecendents can be true or false when consequents are true (i.e., $\{(\top, \top), (\bot, \top)\}$).
Therefore, given
a training set of ground facts $\mathcal{D}_f = \{f_1,\ldots,f_n\}$ assumed to be true, we embed this factual information in \methods by adding it to the logical constraints via a logical conjunction.
E.g., from the example constraint in formula \ref{example_constraint}, if we assume the fact $f_1 = \text{(daffodil, IsAflower)}$ is assumed to be true, we get:
\begin{equation}
    \label{grounded_constraint}
    z_{f_1} = \top, \;\;\;
    \alpha_i' = (z_{f_1} \rightarrow z_{f_2}) \land z_{f_1}
\end{equation}
The assignments satisfying this formula change to $\mu' = \{(\top, \top)\}, \; \mu' \models \alpha_i' \;$, and so does the optimization objective as in equation \ref{applied_semantic_loss_1}.

\section{Experiments}
\label{sec:exps}

We aim to answer these research questions: \textbf{RQ1)} can \methods compete with approaches
using external reasoners without doing so? \textbf{RQ2)} can they require less data than standard fine-tuning?

\textbf{Data.} We evaluate \methods on the BeliefBank data set \citep{kassner2021beliefbank}.
It consists of three pieces:
a ``calibration"  set of $1,072$ annotated facts about $7$ entities of the form \textit{(subject, property, true/false)} used for training,
a ``silver" set of $12,636$ facts about 85 entities used for evaluation,
and a set of $2224$ valid logical implications $\mathcal{D}_c$.
The SL requires defining a set of ground constraints.  We derive these as follows.  For each general constraint in $\mathcal{D}_c$, we lookup the subjects of all facts in the training set: if the antecedent or the consequent fact of the general constraint is known for that subject, we add the subject ground constraint to the dataset.

For RQ1,
we generate two splits: \textit{T1 facts}, appearing either as antecedents or consequents in the constraints; \textit{T2 facts}, appearing exclusively as constraint consequents.
The goal is to correctly guess the consequents by seeing only the antecedents and the constraints. In the calibration set, we count 796 antecedents and 276 consequents, spawning 14,005 grounded constraints. In the silver set, we count 9,504 antecedents and 3,132 consequents, spawning 169,913 grounded constraints.
For RQ2, we split the dataset to test the effects of pure supervised fine-tuning: a portion of random facts from the calibration set is taken with the goal to predict the excluded antecedent or consequent facts.

\textbf{Models.} Following \citet{mitchell2022enhancing}, we work with a pre-trained Macaw-Large model \citep{tafjord2021generalpurpose} (770M parameters) capable of multi-angle question answering with fixed prompt templates.
We adopt the same prompts as in the original paper:  to query for binary beliefs, the model is queried: ``\texttt{\$answer\$ ; \$mcoptions\$ = (A) Yes. (B) No. ; \$question\$ = Is [X] a [Y]?}", and the expected continuation would be: ``\texttt{\$answer\$ = Yes./No.}". Since the training facts are presented in the form \textit{(subject, property)}, we employ the functions used by \citet{mitchell2022enhancing} to fill in natural language templates such as ``\texttt{Is it true that [X] is [Y]?}" or ``\texttt{Is it [X] a [Y]?}".
Further training details are reported in Appendix \ref{appendix_a}.

\textbf{Competitors and Metrics.} We consider two baselines:
the \underline{ConCoRD} logical layer \citep{mitchell2022enhancing} applied to Macaw-Large, using RoBERTa-ANLI \citep{liu2019roberta} for relationship inference, and
a pre-trained \underline{Macaw-Large} model from \cite{tafjord2021generalpurpose} as zero-shot baseline.
In Appendix \ref{inference_time}, we compare our models with the baseline in terms of runtime requirements.
We evaluate our models for \textit{factuality} and \textit{logical self-consistency}.
We measure the former with the $F_1$ score to account for the unbalance between false and true facts
\citep{kassner2021beliefbank}.
Factuality is measured on the two splits (antecedents and consequents) and the complete facts set for both the distributions (calibration, silver).  As in \citep{li2019logicdriven}, we measure \textit{logical self-consistency} as the fraction of non violated constraints  $\alpha_i \in \mathcal{D}_c^{\text{test}}$:
\begin{equation}
\label{eq:log-consistency}
    1 - { |\{ \alpha_i \mid \lnot(z_j \rightarrow z_k)\}|}\ /\  {| \{ \alpha_i \mid z_j \}| }
\end{equation}
i.e., the proportion of unsatisfied formulas where the antecedent is believed to be true.
According to literature, we implicitly refer to ``self-consistency", that is logical consistency based on the model's own beliefs. Logical consistency scores are computed by iterating through the subjects, the general constraints and by querying each belief appearing in the formulas.

\textbf{Results.}  We report factuality and logical consistency scores for the test (Table \ref{table_A}) and training distribution (Table \ref{table_B}). The test distribution involves entities unseen at training time, thus we assume the language model implicitly transfers some semantic knowledge based on similarities in the conceptual space; we further investigate this by calculating pairwise similarities between entity embeddings in Appendix \ref{RDM}.

\begin{table}[!t]
    \centering
    \renewcommand{\arraystretch}{1.2}
        \scriptsize
        \caption{
        \textbf{\methods achieve better logical self-consistency and factuality} as measured via \cref{eq:log-consistency} and $F_1$ scores when compared to classical supervised finetuning (SFT) and baselines using external reasoners such as ConCoRD \citep{mitchell2022enhancing}  measured on test (silver set) facts.
        For RQ1 (\cref{sec:exps}),
        \methods fine-tuned on T1 facts only outperform training-free baseline for all metrics.
        For RQ2, they boost performance in the presence of a small fraction of T1+T2 facts (5-10\%).
        For larger dataset sizes, \methods are competitive for consistency and factuality on consequents.
        A similar trend is visible on training data (\cref{table_B}).
        }
        \label{table_A}
        \begin{center}
            \begin{tabular}{ccccccc}
                \toprule
                        {} & Method & Train Subset & Antecedents $F_1$ & Consequents $F_1$ & Total $F_1$ &  Logical consistency \\
                \midrule
                        \parbox[t]{2mm}{\multirow{4}{*}{\rotatebox[origin=c]{0}{RQ1}}}
                        {} & \multicolumn{1}{l}{ConCoRD} & ~ & ~ & ~ & 0.91 & 0.91 \\
                        {} & \multicolumn{1}{l}{Macaw-Large} & ~ & 0.52 & 0.90 & 0.81 & 0.83 \\
                        {} & \multicolumn{1}{l}{SFT} & T1 & 0.13 & 0.01 & 0.03 & 0.72 \\
                        {} & \multicolumn{1}{l}{\method} & T1 & \textbf{0.79} & \textbf{0.98} & \textbf{0.96} & \textbf{0.99} \\
                \midrule
                        \parbox[t]{2mm}{\multirow{6}{*}{\rotatebox[origin=c]{0}{RQ2}}}
                        {} & \multicolumn{1}{l}{SFT} & T1+T2 (5\%) & 0.23 & 0.78 & 0.72 & 0.82 \\
                        {} & \multicolumn{1}{l}{\method} & T1+T2 (5\%) & \textbf{0.67} & \textbf{0.83} & \textbf{0.81} & \textbf{0.92} \\
                \cmidrule(l){2-7}
                        {} & \multicolumn{1}{l}{SFT} & T1+T2 (10\%) & \textbf{0.55} & \textbf{0.97} & \textbf{0.91} & 0.90 \\
                        {} & \multicolumn{1}{l}{\method} & T1+T2 (10\%) & 0.45 & \textbf{0.97} & 0.89 & \textbf{0.93} \\
                \cmidrule(l){2-7}
                        {} & \multicolumn{1}{l}{SFT} & T1+T2 (75\%) & \textbf{0.85} & \textbf{0.99} & \textbf{0.97} & \textbf{0.98} \\
                        {} & \multicolumn{1}{l}{\method} & T1+T2 (75\%) & 0.79 & \textbf{0.99} & 0.95 & \textbf{0.98} \\
                \bottomrule
            \end{tabular}
        \end{center}
\end{table}

We firstly observe a net improvement in both factuality and logical consistency with semantic loss fine-tuning, compared to pre-trained Macaw-Large and the ConCoRD variant (Table \ref{table_A}, RQ1). Standard supervised fine-tuning on antecedent facts is insufficient: due to a class imbalance between true facts ($\sim{10\%}$) and false facts ($\sim{90\%}$), the model tends to label any statement as ``false"; moreover, no knowledge about consequent facts is introduced. This is accentuated in the training distribution (Table \ref{table_B}): with semantic loss, the model attains higher factuality and logical consistency in both the standard setup and with a small amount of cheating data. Finally, assuming sufficient domain knowledge overlap (Appendix \ref{RDM}), high logical consistency scores on the test distribution suggest the imposed logical structures in the conceptual space can be generalized.

Assuming the language model can access to a portion of consequent facts (\cref{table_A}, RQ2), semantic loss fine-tuned \methods still yields better logical consistency and factuality for unseen consequents in low-data regimes (e.g. 5-10\% of the T1+T2 dataset) compared to canonical supervised fine-tuning.
When they are allowed to see more data (e.g. 75\% of the T1+T2 dataset), traditionally finetuned models can ``cheat" and directly learn about the consequents (somehow equivalent to memorizing a single row of the truth table).
In this scenario, \methods achieve compareble logical self-consistency and factuality over consequents, but not on the antecedents.

\section{Conclusion and Further Work}

Our results show that \methods improve upon ConCoRD in terms of factuality and self-consistency in complex reasoning tasks, especially when queried on unseen facts.  This suggests that probabilistic reasoning objectives can impose structure in a language model's conceptual space.
In future work, we aim to further investigate the transfer of logical structures across entities with semantic relations, such as e.g. abstraction or homology. We also consider the implications of scaling language models, in terms of logical consistency and training efficiency.
Finally, we plan to extend our analysis to more complex logical operators and reasoning scenarios \citep{vergari2021compositional} that can support provably  consistent reasoning \citep{ahmed2022semantic}.

\subsubsection*{Acknowledgments}

Funded by the European Union. Views and opinions expressed are however those of the author(s) only and do not necessarily reflect those of the European Union or the European Health and Digital Executive Agency (HaDEA). Neither the European Union nor the granting authority can be held responsible for them. Grant Agreement no. 101120763 - TANGO.
AV is supported by the "UNREAL: Unified Reasoning Layer for Trustworthy ML" project (EP/Y023838/1) selected by the ERC and funded by UKRI EPSRC.

\clearpage

\bibliography{iclr2024_conference}
\bibliographystyle{iclr2024_conference}

\clearpage
\appendix

\section{Training details}
\label{appendix_a}

We fine-tune our models for $5$ epochs keeping the learning rate fixed to $\gamma = 3\cdot10^{-4}$ on $1$--$2$ nVidia A30 GPUs.  Each model took approximately $35$ minutes to train.  We use AdamW \citep{loshchilov2016sgdr} as optimizer with a default weight decay $\lambda = 10^{-2}$. 

\section{Evaluation details} \label{appendix_b}

\subsection{Evaluation on the training distribution}
We implement the same evaluation functions from the original ConCoRD codebase (\href{https://github.com/eric-mitchell/concord/}{\texttt{https://github.com/eric-mitchell/concord/}}). For ConCoRD, we report  only the scores computed with the original code and the provided cache inference facts.

\begin{table}[htbp]
    \centering
    \renewcommand{\arraystretch}{1.2}
    \scriptsize
    \caption{
        \textbf{\methods achieve better logical self-consistency and
        factuality} as measured via \cref{eq:log-consistency} and $F_1$ scores
        when compared to classical supervised finetuning (SFT) and baselines
        using external reasoners such as ConCoRD \citep{mitchell2022enhancing}
        measured on train (calibration set) facts.
        For RQ1 (\cref{sec:exps}), \methods fine-tuned on T1 facts only
        outperform training-free baseline for all metrics.
        For RQ2, they boost performance in the presence of a small fraction of
        T1+T2 facts (5-10\%).
        For larger dataset sizes, \methods are competitive for consistency and
        factuality on consequents.
    }
    \label{table_B}
    \begin{center}
        \begin{tabular}{ccccccc}
            \toprule
                    {} & Method & Train size & Antecedents $F_1$ & Consequents $F_1$ & Total $F_1$ & Logical consistency \\
            \hline
                    \parbox[t]{2mm}{\multirow{4}{*}{\rotatebox[origin=c]{0}{RQ1}}}
                    {} & \multicolumn{1}{l}{ConCoRD} & {} & {} & {} & 0.91 & 0.91 \\
                    {} & \multicolumn{1}{l}{Macaw-Large} & {} & 0.47 & 0.84 & 0.78 & 0.82 \\
                    {} & \multicolumn{1}{l}{SFT} & T1 & 0.46 & 0.08 & 0.14 & 0.79 \\
                    {} & \multicolumn{1}{l}{\method} & T1 & \textbf{0.98} & \textbf{0.99} & \textbf{0.99} & \textbf{1.00} \\
            \midrule
                    \parbox[t]{2mm}{\multirow{6}{*}{\rotatebox[origin=c]{0}{RQ2}}}
                    {} & \multicolumn{1}{l}{SFT} & T1+T2 (5\%) & 0.31 & 0.73 & 0.69 & 0.90 \\
                    {} & \multicolumn{1}{l}{\method} & T1+T2 (5\%) & \textbf{0.34} & \textbf{0.77} & \textbf{0.72} & \textbf{0.92} \\
            \cmidrule(l){2-7}
                    {} & \multicolumn{1}{l}{SFT} & T1+T2 (10\%) & 0.48 & 0.88 & 0.85 & 0.87 \\
                    {} & \multicolumn{1}{l}{\method} & T1+T2 (10\%) & \textbf{0.52} & \textbf{0.95} & \textbf{0.89} & \textbf{0.91} \\
            \cmidrule(l){2-7}
                    {} & \multicolumn{1}{l}{SFT} & T1+T2 (75\%) & \textbf{0.69} & \textbf{1.00} &\textbf{0.97} & 0.97 \\
                    {} & \multicolumn{1}{l}{\method} & T1+T2 (75\%) & 0.65 & \textbf{1.00} & \textbf{0.97} & \textbf{0.99} \\
            \bottomrule
        \end{tabular}
    \end{center}
\end{table}

\subsection{Temporal efficiency Analysis}
\label{inference_time}

\begin{table}[htbp!]
    \centering
    \renewcommand{\arraystretch}{1.2}
    \small
    \caption{\textbf{\methods require much less time for inference at the price of a single training step. } We report time elapsed in seconds for (1) training (calibration set only) our model; (2) inference (calibration + silver set) with our model or with ConCoRD+RoBERTa-ANLI. For ConCoRD, we sum time elapsed for QA inference, NLI inference and running the solver. All runs are measured on a single NVIDIA A30.}
    \begin{center}
        \begin{tabular}{ccc}
            \toprule
                    Method & Training time (s) & Inference time (s) \\
            \hline
                    ConCoRD & - & 3669.33 \\
                    \method & 2124.48 & \textbf{2405.28} \\
            \bottomrule
        \end{tabular}
    \end{center}
\end{table}

\subsection{Semantic overlap between train and test}
\label{RDM}

We measure the semantic overlap between the training and test distribution by constructing a Representation Dissimilarity Matrix (RDM) of Macaw's embeddings (token average) between training and test entities. The main assumption is that semantically similar subjects may have similar properties, as a proxy for domain knowledge transfer. \\

\begin{figure}[h]
    \begin{center}
        \hspace*{-1.9in}
        \includegraphics[width=1.8\textwidth]{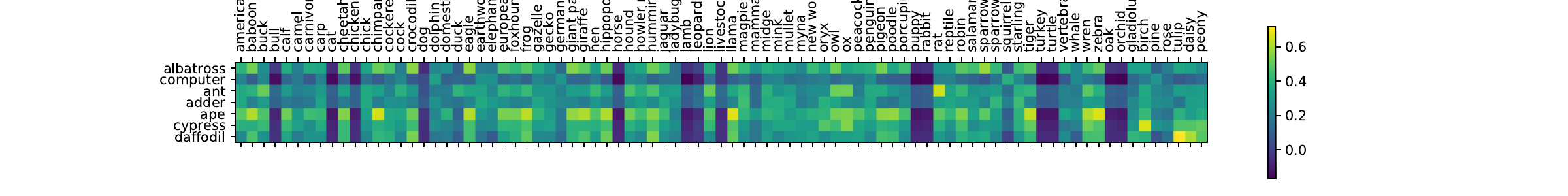}
    \end{center}
    \caption{Pairwise cosine similarities between entities in the training distribution (calibration, rows) and the test distribution (silver, columns).}
\end{figure}

\end{document}